%% file: main.tex
\documentclass[11pt,a4paper]{article}
\usepackage[hyperref]{emnlp2019}
\usepackage{times}
\usepackage{latexsym}
\usepackage{algorithm}
\usepackage[noend]{algpseudocode}
\usepackage{url}
\usepackage{stackengine}
\setstackEOL{\\}
\usepackage{amsmath}
\usepackage{amssymb}
\usepackage{subcaption}
\usepackage{multirow}
\usepackage{multicol}
\usepackage{color}
\usepackage{graphicx}

\aclfinalcopy 

\title{PullNet: Open Domain Question Answering with \\Iterative Retrieval on Knowledge Bases and Text}

\author{Haitian Sun ~~ Tania Bedrax-Weiss ~~ William W. Cohen \\
    Google AI Language \\
   \texttt{\{haitiansun,tbedrax,wcohen\}@google.com} \\
  }

\date{}

\begin{document}
\maketitle

\input{00_abstract.tex}
\input{01_introduction.tex}
\input{02_related_work.tex}

\input{03_method.tex}
\input{04_experiments.tex}
\input{05_conclusion.tex}

\input{emnlp2019.bbl}
\bibliographystyle{acl_natbib}

\end{document}

%% file: 00_abstract.tex
\begin{abstract}
We consider open-domain queston answering (QA) where answers are drawn from either a corpus, a knowledge base (KB), or a combination of both of these.  We focus on a  setting in which a corpus is supplemented with a large but incomplete KB, and on questions that require non-trivial (e.g., ``multi-hop'') reasoning.  We describe PullNet, an integrated framework for (1) learning what to retrieve (from the KB and/or corpus) and (2) reasoning with this heterogeneous information to find the best answer. PullNet uses an {iterative} process to construct a question-specific subgraph that contains information relevant to the question. In each iteration, a graph convolutional network (graph CNN) is used to identify subgraph nodes that should be expanded using retrieval (or ``pull'') operations on the corpus and/or KB.  After the subgraph is complete, a similar graph CNN is used to extract the answer from the subgraph.  This retrieve-and-reason process allows us to answer multi-hop questions using large KBs and corpora.  PullNet is weakly supervised, requiring question-answer pairs but not gold inference paths.  Experimentally PullNet improves over the prior state-of-the art, and in the setting where a corpus is used with incomplete KB these improvements are often dramatic.  PullNet is also often superior to prior systems in a KB-only setting or a text-only setting.
\end{abstract}

%% file: 01_introduction.tex
\section{Introduction}

Open domain Question Answering (QA) is the task of finding answers to questions posed in natural language, usually using text from a corpus  \cite{dhingra2017quasar,JoshiTriviaQA2017, dunn2017searchqa}, or triples from a knowledge base (KB) \cite{zelle1996learning,zettlemoyer2005learning,webqsp}.  Both of these approaches have limitations. Even the largest KBs are incomplete \cite{Min2013}, which limits recall of a KB-based QA system.  On the other hand, while a large corpus may contain more answers than a KB, the diversity of natural language makes corpus-based QA difficult \cite{chen2017reading, welbl2018constructing,kwaitkowski2019,yang2018hotpotqa}.

In this paper we follow previous research \cite{sawant2019neural,graftnet} in deriving answers using both a corpus and a KB.  We focus on tasks in which questions require compositional (sometimes called ``multi-hop'') reasoning, and a setting in which the KB is incomplete, and hence must be supplemented with information extracted from text.  We also restrict ourselves in this paper to answers which correspond to KB entities.  For this setting, we propose an integrated framework for (1) \emph{learning what to retrieve}, from either a corpus, a KB, or a combination, and (2) \emph{combining this heterogeneous information} into a single data structure that allows the system to reason and find the best answer.  In prior work, this approach was termed an \textit{early fusion} approach, and shown to improve over \textit{late fusion} methods, in which two QA systems, one corpus-based and one KB-based, are combined in an ensemble.  

The system we describe, PullNet, builds on the GRAFT-Net\footnote{Graphs of Relations Among
Facts and Text Networks.} early fusion system.  GRAFT-Net uses heuristics to build a question-specific subgraph which contains sentences from the corpus, entities from the KB, and facts from the KB.  A graph CNN \cite{kipf2016semi, li2015gated, schlichtkrull2017modeling} variant is then used to reason over this graph and select an answer.  However, as we will show experimentally, GRAFT-Net's heuristics often produce subgraphs that far from optimal: they are generally much larger than necessary, and often do not contain the answer at all.

PullNet also uses a reasoning process based on a graph CNN to find answers.  However, PullNet \textit{learns} how to construct the subgraph, rather than using an \textit{ad hoc} subgraph-building strategy.  More specifically, PullNet relies on a small set of retrieval operations, each of which takes a node in an existing subgraph, and then expand the node by retrieving new information from the KB or the corpus.  PullNet learns when and where to apply these ``pull'' operations with another graph CNN classifier.   The ``pull'' classifier is weakly supervised , using question-answer pairs for supervision.  

The end result is a learned {iterative} process for subgraph construction, which begins with a small subgraph containing only the question text and the entities which it contains, and gradually expands the subgraph to contain information from the KB and corpus that are likely to be useful.  The incremental question-guided subgraph construction process results in high-recall subgraphs that are much smaller than the ones created heuristically, making the final answer extraction process easier.  The process is especially effective for multi-hop questions, which naively would require expanding the subgraph to include all corpus and KB elements that are $k$ hops away from the question.

PullNet improves over the current state-of-the-art for KB-only QA on several benchmark datasets, and is superior to, or competitive with, corpus-only QA on several others.  For multi-hop questions, this improvement is often dramatic: for instance, MetaQA \cite{metaqa} contains multi-hop questions based on a small movie KB, originally associated with the WikiMovies dataset \cite{miller2016key}.  In a KB-only setting, PullNet improves hits-at-one performance for 3-hop MetaQA questions from 62.5\%  to 91.4\%.  Perhaps more interestingly, PullNet obtains performance of 85.2\% hits-at-one with a KB from which half of the triples have been removed, if that KB is supplemented with a corpus.  We note that this result improves by 7\% (absolute improvement) over a pure corpus-based QA system,  and  by more than 25\% over a pure KB-based QA system. In a similar incomplete-KB setting, PullNet improves over GRAFTNet by 6.8\% on the ComplexWebQuestions dataset \cite{complexwebq}.

%% file: 02_related_work.tex
\section{Related Work}


This paper has focused on QA for multi-hop questions using large KBs and text corpora as the information sources from which answers can be drawn. The main technical contribution is an iterative question-guided retrieval mechanism that retrieves information from KBs, corpora, or combinations of both.  The iterative retrieval makes it possible to follow long paths of reasoning on large KBs, and as we show below, this leads to new state-of-the art results on two datasets of this sort.  The hybrid KB/text retrieval, and the ability to reason over combinations of text and KBs, lets us extend these results to settings where only text is available, and to settings where text and an incomplete KB is available.  These contributions are based on much prior work.  

A long line of QA models have been developed which answer questions based on a single passage of text \cite{dhingra2016gated, yu2018qanet, seo2016bidirectional, gao2018dynamic, liu2017phase,devlin2018bert}. Generally, these ``reading comprehension'' systems operate by encoding the passage and question into an embedding space, and due to memory limitations cannot be applied to a large corpus instead of a short passage.  To address this limitation a number of systems have been designed which use a ``retrieve and read'' pipeline 
\cite{chen2017reading, dhingra2017quasar, JoshiTriviaQA2017,dunn2017searchqa,wang2018r, wang2017evidence}, in which a retrieval system with high recall is piped into a reading comprehension system that can find the answer.  An alternative approach is ``phrase-indexed'' QA \cite{seo2018phrase}, where embedded phrases in a document are indexed and searched over.  Such systems differ from PullNet in that only a single round of retrieval is used; however, for questions that requires multi-hop reasoning, we believe is difficult for a single retrieval step to find the relevant information.  These systems are also not able to use both KB and text for QA.  

SplitQA \citep{complexwebq} is a text-based QA system that cam decompose complex questions (e.g. conjunction, composition) into simple subquestions, and perform retrieval on them sequentially. Although it uses iterative retrieval for multi-hop questions, unlike PullNet, SplitQA  does not also use a KB as an information source. Also, since SplitQA has been applied only to the Complex WebQuestions dataset, it is unclear how general its question decomposition methods are.

There has also been much work on QA from KBs alone, often using methods based on memory networks \cite{Sukhbaatar2015EndToEndMN}, semantic parsing \cite{zelle1996learning,zettlemoyer2005learning} or reinforcement learning \citet{das2017go}. Extending such KB-based work to also use text, however, is non-trivial. In \cite{DBLP:journals/corr/DasZRM17} one approach for a hybrid text-based and KB-based QA system is described, where key-value memory networks are used to store text as well as KB facts encoded with a universal schema representation \cite{riedel2013relation}.  We use a similar method as one of our baselines, following \cite{graftnet}. Another line of QA work from text and KBs is exemplified by AQQU \cite{bast2015more}  and its successors \cite{sawant2019neural}.  These systems focus on questions that, like the questions in the SimpleWebQuestions dataset, can be interpreted as identifying an entity based on a relationship and related entity, plus (potentially) additional restrictions described in text, and it is unclear how to extend such approaches to multi-hop questions.  

GRAFT-Net \cite{graftnet} supports multi-hop reasoning on both KBs and text by introducing a question subgraph built with facts and text, and uses a learned graph representation \cite{kipf2016semi, li2015gated, schlichtkrull2017modeling, scarselli2009graph} to perform the ``reasoning'' required to select the answer.  We use the same representation and reasoning scheme as GRAFT-Net, but do not require that the entire graph be retrieved in a single step.  In our experimental comparisons, this gives significant performance gains for multi-hop reasoning tasks.

Combinations of KBs and text have also been used for relation extraction and Knowledge Base Completion (KBC) \cite{lao2012reading, toutanova2015representing, das2017go}. The QA task differs from KBC in that in QA, the inference process must be conditioned on a natural-language question, which leads to different constraints on which methods can be used.

%% file: 03_method.tex
\section{The PullNet Model}
PullNet retrieves from two ``knowledge sources'', a text corpus and a KB. Given a question, PullNet will use these to construct a \textit{question subgraph} that can be used to answer the question.  The question subgraph is constructed iteratively.  Initially the subgraph depends only on the question.  PullNet then iteratively expands the subgraph by choosing nodes from which to ``pull'' information about, from the KB or corpus as appropriate.  The question subgraph is heterogeneous, and contains both entities, KB triples, and entity-linked text.

In this section, we will first introduce notation defining the heterogeneous graph structure we use. Then we will introduce the general iterative retrieval process. Finally, we will discuss the retrieval operations used on the corpus and KB, and the classification operations on the graph which determine where to perform the retrievals.

\subsection{The Question Subgraph} \label{sec:notation}

A \textit{question subgraph} for question $q$, denoted $\mathcal{G}_q = \{\mathcal{V}, \mathcal{E}\}$,  is a hetogeneous graph that contains information from both the text corpus and the KB relevant to $q$. Let $\mathcal{V}$ denote the set of vertices, which we also call nodes. Following GRAFT-Net \cite{graftnet}, there are three types of nodes: entity nodes $\mathcal{V}_e$, text nodes $\mathcal{V}_d$, and fact nodes $\mathcal{V}_f$, with $\mathcal{V} = \mathcal{V}_e \cup \mathcal{V}_d \cup \mathcal{V}_f$. A entity node $v_e \in \mathcal{V}_e$ represents an entity from the knowledge base. A text node $v_d \in \mathcal{V}_d$ represents a document from the corpus, with a sequence of tokens denoted $(w_1, \dots, w_{|d|})$. In this paper, a document is always a single sentence, to which an entity linker \cite{ji2014overview} has been applied to detect and ground entity mentions.
A fact node $v_f \in \mathcal{V}_f$ represents a triplet $(v_s, r, v_o)$ from the KB, with subject and objects $v_s, v_o \in \mathcal{V}_e$ and relation $r$. We let $\mathcal{E}$ denote the set of edges between nodes. An edge connects a fact node $v_f$ and an entity node $v_e$ exists iff fact $v_f$ has $v_e$ as its subject or object. An edge connects a text node $v_d$ with entity node $v_e$ iff the entity is mentioned in the text.


\subsection{Iterative subgraph expansion and classification}

\subsubsection{Overview}

We start with a question subgraph $\mathcal{G}_q^0 = \{\mathcal{V}^0, \mathcal{E}^0\}$ where $\mathcal{V}^0 = \{e_{q_i}\}$ is the list of entities mentioned in the question and $\mathcal{E}^0$ is an empty set. We iteratively expand the question subgraph $\mathcal{G}_q^0$ until it contains the information required to answer the question. 

The algorithm is shown in Alg \ref{alg:pullnet}.  Briefly, we expand the graph in $T$ iterations.  In each iteration, we choose $k$ entities to expand, and then for each selected entity, we retrieve a set of related documents, and also a set of related facts.  The new documents are then passed through an entity-linking system to identify entities that occur in them, and the head and tail entities of each fact will also be extracted.  The last stage in the iteration is to update the question graph by adding all these new edges.  After the $t$-th iteration of expansion, an additional classification step is applied to the final question subgraph which predicts the answer entity.

\begin{algorithm}[H]
\small
\caption{PullNet}
\begin{algorithmic}[1]
\State Initialize question graph $G_q^0$ with question $q$ and question entities, with $\mathcal{V}^0 = \{e_{q_i}\}$ and $\mathcal{E}^0 = \emptyset$.
\For{$t = 1, \cdots, T $}
	\State \Longunderstack[l]{Classify and select the entity nodes in the graph with \\ probability larger than $\epsilon$}
	\vspace{-2pt}$$\{v_{e_i}\} = \texttt{classify\_pullnodes($G_q^{t},k)$}$$\vspace{-12pt}
    \ForAll {$v_e$ in $\{v_{e_i}\}$}
    	\State Perform pull operation on selected entity nodes
    	\vspace{-8pt}
    	\begin{align*}
    	    \{v_{d_i}\} = \texttt{pull\_docs($v_e, q)$}\\
    	    \{v_{f_i}\} = \texttt{pull\_facts($v_e, q$)}
    	\end{align*}\vspace{-16pt}
    	
    	\ForAll {$v_d$ in $\{v_{d_i}\}$}
    	    \State Extracted entities in new document nodes
    	        \vspace{-8pt}$$\{v_{e(d)_i}\} = \texttt{pull\_entities($v_d$)}$$\vspace{-16pt}
    	\EndFor
    	\ForAll {$v_f$ in $\{v_{f_i}\}$}
    	    \State Extract head and tail of new fact nodes  \vspace{-8pt}$$\{v_{e(f)_i}\} = \texttt{pull\_headtail($v_f$)}$$\vspace{-16pt}
    	\EndFor
    \EndFor 
    \State Add new nodes and edges to question graph 
        \vspace{-8pt}$$G_q^{t + 1} = \texttt{update($G_q^{t}$)} $$\vspace{-16pt}
\EndFor
\State Select entity node in final graph that is the best answer
\vspace{-8pt}$$v_{\textnormal{ans}} = \texttt{classify\_answer($G_q^{T})$}$$\vspace{-16pt}
\end{algorithmic}
\label{alg:pullnet}
\end{algorithm}

We will now describe the operations used in Alg \ref{alg:pullnet}. 

\subsubsection{Pull Operations}

Pull operations either retrieve information from a knowledge source, or extract entities from a fact or document.

The two extraction operations are relatively simple.  The \texttt{pull\_entities($v_d$)} operation inputs a document node $v_d$, calls an entity linker and returns all entities mentioned in $v_d$.  The \texttt{pull\_headtail($v_f$)} operation inputs a fact node $v_f$ and returns the subject and object entity of fact $v_f$.

The retrieval operations are more complex. 
The \texttt{pull\_docs($v_e, q)$)} 
operation retrieves relevant documents from the corpus. We use an IDF-based retrieval system, Lucene \cite{mccandless2010lucene} and  assume that all sentences have been entity-linked prior to being indexed. 
The retrieved documents are constrained to link to entity $v_e$, and are ranked by their IDF similarity to the question $q$.  Only the top $N_d$ documents in this ranking are returned.  

The \texttt{pull\_facts($v_e, q$)} operation retrieves the top $N_f$ facts from the KB about entity $v_e$. The retrieved facts are constrained to have $v_e$ as their subject or object, and are ranked based on the similarity  $S(r, q)$ between the fact's relation $r$ and the question $q$.  Since it is not obvious how to assess relevance of a fact to a question $q$, we learn a $S(r, q)$ as follows. Let $h_r$ be an embedding of relation $r$, which is looked up from an embedding table, and let $q = (w_1, \dots, w_{|q|})$ be the sequence of words for question $q$.  Similarity is defined as the dot-product of the last-state LSTM representation for $q$ with the embedding for $r$.  This dot-product is then passed through a sigmoid function to bring it into a range of $[0,1]$: as we explain below, we will train this similarity function as a classifier which predicts which retrieved facts are relevant to the question $q$.  The final ranking method for facts is thus
\begin{align*}
&h_q = \text{LSTM}(w_1, \dots, w_{|q|} ) \in \mathbb{R}^n \\
&S(r, q) = \text{sigmoid} (h_r^T~h_q)
\end{align*}

\subsubsection{Classify Operations}
Two types of \texttt{classify} operations are applied to the nodes in a subgraph $G_q^{t}$.  These operations are applied only to the entity nodes in the graph, but they are based on node representations computed by the graph CNN, so the non-entity nodes and edges also affect the classification results.

During subgraph construction, the \texttt{classify\_pullnodes($G_q^{t}$)} operation returns a probability an entity node $v_e$ should be expanded in the next iteration. In subgraph construction, we choose the $k$ nodes with the highest probability in each iteration. 
After the subgraph is complete, the
\texttt{classify\_answer($G_q^{t}$)} operation predicts whether an entity node answers the question.  The highest-scoring entity node is returned as the final answer.  

We use the same CNN architecture used by GRAFT-Net \cite{graftnet} for classification. GRAFT-Net's source is open-domain and it supports node classification on heterogeneous graphs containing facts, entities, and documents.  Graft-Net differs from other graph CNN implementations, in using special mechanisms to distribute representations across different types of nodes and edges: notably, document nodes are represented with LSTM encodings, extended by mechanisms that allow the representations for entity nodes $v_e$ to be passed into a document $v_d$ that mentions $v_e$, and mechanisms that allow the LSTM hidden states associated with an entity mention to be passed out of $v_d$ to the associated entity node $v_e$.

\subsubsection{The Update Operation}
The \texttt{update} operation takes the question subgraph $G_q^{t-1}$ from the previous iteration and updates it by adding the newly retrieved entity nodes $\{v_{e(f)_i}\} \cup \{v_{e(d)_i}\}$, the text nodes $\{v_{d_i}\}$, and the fact nodes $\{v_{f_i}\}$. It also updates the set of edges $\mathcal{E}$ based on the definitions of Section~\ref{sec:notation}. Note that some new edges are derived when pull operations are performed on text nodes and fact node, but other new edges may connect newly-added nodes with nodes that already exist in the previous subgraph.

\subsection{Training}

To train PullNet, we assume that we only observe question and answer pairs, i.e., the actual inference chain required to answer the question is latent. We thus need to use weak supervision to train the classifiers and similarity scores described above.

To train these models, we form an approximation of the ideal question subgraph for question $q$ as follows.  Note that in training, the answer entities are available.  We use these to find all shortest paths in the KB between the question entities and answer entities.  Each entity $e$ that appears in such a shortest path will be marked as a \textit{candidate intermediate entities}.  For each candidate intermediate entity $e$ we record its minimal distance $t_e$ from the question nodes.

When we train the \texttt{classify\_pullnodes} classifier in iteration $t$, we treat as positive examples only those entities $e'$ that are connected to a candidate intermediate entity $e$ with distance $e_t=t+1$.  Likewise in training the similarity function $S(h_r,q)$ we treat as positive relations leading to candidate intermediate entities $e$ at distance $e_t=t+1$.  This encourages the retrieval to focus on nodes that lie on shortest paths to an answer.

In training we use a variant of teacher forcing.  When training, we pull from all entity nodes with a predicted score larger than some threshold $\epsilon$, rather than only the top $k$ nodes.  If, during training, a candidate intermediate entity is not retrieved in iteration $t_e$,  we add it to the graph anyway. The values $T$ and $\epsilon$ are hyperparameters, but here we always pick for $T$, the number of retrieval steps, maximum length of the inference chain needed to ensure full coverage of the answers.  

We use the same classifier on the graph in the retrieval step as in answer selection, except that we change last fully-connected layer.  The classifiers used for retrieval in the different iterations are identical.

The learned parts of the model are implemented in PyTorch, using an ADAM optimizer \cite{kingma2014adam}, and the full retrieval process of Alg \ref{alg:pullnet} is performed on each minibatch.

%% file: 04_experiments.tex
\section{Experiments and Results}
\subsection{Datasets}
\noindent \textbf{MetaQA} \cite{metaqa} consists of more than 400k single and multi-hop (up to 3-hop) questions in the domain of movies. The questions were constructed using the knowledge base provided by the WikiMovies \cite{miller2016key} dataset.  We use the ``vanilla'' version of the queries: for this version, the 1-hop questions in MetaQA are exactly the same as WikiMovies. We use the KB and text corpus that are supplied with the WikiMovies dataset \cite{miller2016key}, and run use a simple exact match on surface forms to perform entity linking.
The KB used here is relatively small, with about 43k entities and 135k triples.

\noindent \textbf{WebQuestionsSP} \cite{webqsp} contains 4737 natural language questions that are answerable using Freebase\footnote{We use the same train/dev/test splits as GRAFT-Net \cite{graftnet}.}. The questions require up to 2-hop reasoning from knowledge base, and 1-hop reasoning using the corpus. We use Freebase as our knowledge base but for ease of experimentation restrict it to a subset of Freebase which contains all facts that are within 2-hops of any entity mentioned in the questions of WebQuestionsSP. This smaller KB contains 164.6 million facts and 24.9 million entities.  We use Wikipedia as our text corpus and again use a simple entity-linker: we link entities by exact matching to any surface form annotated in FACC1 project \cite{facc1}.\footnote{If two overlapping spans are possible matches, we match only the longer of them.}

\noindent \textbf{Complex WebQuestions 1.1 (Complex WebQ)} \cite{complexwebq} is generated from Web\-QuestionsSP by extending the question entities or adding constraints to answers, in order to construct more complex multi-hop  questions\footnote{We use Complex WebQuestions v1.1, where the author re-partition the train/dev/test data to prevent the leakage of information.}. There are four types of question: composition (45\%), conjunction (45\%), comparative (5\%), and superlative (5\%). The questions require up to 4-hops of reasoning on knowledge base and 2-hops on the corpus. We use the same 
KB and corpus as used for WebQuestionsSP.

\begin{table}[htp]
\small
\centering
\begin{tabular}{lccc}
\hline
               & Train   & Dev    & Test   \\
\hline
MetaQA 1-hop   & 96,106  & 9,992  & 9,947  \\
MetaQA 2-hop   & 118,980 & 14,872 & 14,872 \\
MetaQA 3-hop   & 114,196 & 14,274 & 14,274 \\
WebQuestionsSP & 2,848   & 250    & 1,639  \\
Complex WebQ   & 27,623  & 3,518  & 3,531 \\
\hline
\end{tabular}
\caption{Statistics of all datasets.}
\end{table}

\subsection{Tasks}

We explored several different QA settings:  KB only (complete), corpus only, incomplete KB only, and incomplete KB paired with the corpus.

In the \textit{KB only complete} setting, the answer always exists in knowledge base: for all of these datasets, this is true because the questions were crowd-sourced to enforce this conditions.  This is the easiest setting for QA, but arguably unrealistic, since with a more natural distribution of questions, a KB is likely to be incomplete.  Note that PullNet can be used in this setting by simply removing the \texttt{pull\_docs} operation.

In the \textit{text only} setting we use only the corpus.  Again, PullNet can operate in this setting by removing the \texttt{pull\_facts} operation.

In the \textit{incomplete KB} setting, we simulate KB-based QA on anincomplete KB by randomly discarding some of the triples in the KB: specifically, we randomly drop a fact from the knowledge base with  probability $p=50\%$.

In the \textit{incomplete KB plus text setting}, we pair the same incomplete knowledge base with the corpus.  In principle this allows a learned QA system to adopt many different hybrid strategies.  For simple 1-hop queries, a model might use the KB when the required fact exists, and ``back off'' to text when the KB is missing information.  In multihop or conjunctive queries, the reasoning done in answering the question might involve combining inferences done with text and inferences done with KB triples.

\textbf{Comment.} One point to note is that our training procedure is based on finding shortest paths in a complete KB, and we use this same procedure on the incomplete KB setting as well.  Thus the weak training that we use should, in the incomplete-KB settings, be viewed as a form of weak supervision, with labels that are intermediate in informativeness between pure distant training (with only question-answer pairs) and gold inference paths (a setting that has been extensively investigated on some of these datasets, in particular Web\-Questions\-SP).


\subsection{Baselines}
We choose Key-Value Memory Network \cite{miller2016key} and GRAFT-Net \cite{graftnet} as our baseline models: to the best of our knowledge, these are the only ones that can use both text and KBs for question answering. However, both models are limited in the number of facts and text that can fit into memory. Thus, we create a separate retrieval process as a pre-processing step, which will be discussed below.

\textbf{Key-Value Memory Network (KVMem)} \cite{miller2016key} maintains a memory table which stores KB facts and text encoded into key-value pairs. The encoding of KB facts is the same as is presented in \citet{miller2016key}. For text, we use an bi-directional LSTM to encode the text for the key and take the entities mentioned in the text as values. Our implementation shows comparable performance on the WikiMovies (MetaQA 1-hop) datasets as reported previously, as shown in Table \ref{tbl:metaqa_results}.

For \textbf{GRAFT-Net} \cite{graftnet}, we use the implementation published by the author\footnote{\url{https://github.com/OceanskySun/GraftNet}}; however, we run it on somewhat different data as described in Section \ref{sec:subgraph_retrieval}.

\subsection{Subgraph Retrieval for Baseline Models} \label{sec:subgraph_retrieval}
\input{metaqa_results.tex}

\textbf{Text} was retrieved (non-iteratively) using the IDF-based similarity to the question. It is less obvious how to perform KB retrieval: we would like to retrieve as many facts as possible to maximize the recall of answers, but it is infeasible to take all facts that are within $k$-hop away from question entities since the number grows exponentially, and there is no widely-used heuristic, corresponding to IDF-based text retrieval, for extracting subsets of a KB.  We elected to first run the PageRank-Nibble algorithm \cite{apr} with $\epsilon=1e^{-6}$ and pick the top $m$ entities.  PageRank-Nibble efficiently approximates a personalized PageRank (aka random walk with reset) seeded from the questions (not the answers, which are of course not available at test time).  We then find all entities from these top $m$ entities that are within $k$-hops of the question entities. Finally, we collect all facts from the knowledge base that connect any pair of the retrieved entities. This allows us to easily vary the number of KB facts that are retrieved. 

The retrieval results are shown in Table \ref{tbl:metaqa_retrieval} and \ref{tbl:webq_retrieval}.  For the smaller MetaQA KB, the proposed retrieval method finds high-coverage graphs.  For ComplexWebQuestions, the recall is only 64\%, even with a graph with nearly 2000 nodes: this is expected, since retrieving the relevant entities for a multi-hop question from a KB with millions of entities is a difficult task.

\begin{table}[htp]
\centering
\small
\begin{tabular}{lcc}
\hline
             & wikimovies-KB & wikimovies-KB (50\%) \\ \hline
1-hop & 0.995 / 9.17         & 0.544 / 4.58                \\ \hline
2-hop & 0.983 / 47.3         & 0.344 / 28.6                \\ \hline
3-hop & 0.923 / 459.2        & 0.522 / 316.6               \\ \hline
\end{tabular}
\caption{Retrieval performance for baseline retrieval model on MetaQA with 500 PageRank-Nibble entities (recall / \# entities in graph)\label{tbl:metaqa_retrieval}}
\end{table}
\vspace{-12pt}
\begin{table}[htp]
\centering
\small
\begin{tabular}{lcc}
\hline
             & Freebase & Freebase (50\%) \\ \hline
WebQuestionsSP & 0.927 / 1876.9         & 0.485 / 1212.5                \\ \hline
Complex WebQ & 0.644 / 1948.7        & 0.542 / 1849.2               \\ \hline
\end{tabular}
\caption{Retrieval performance for baseline retrieval model on WebQuestionsSP and Complex WebQuestions with 2000 PageRank-Nibble entities (recall / \# entities in graph)\label{tbl:webq_retrieval}}
\vspace{-12pt}
\end{table}

\subsection{Main Results}
\subsubsection{MetaQA}

The experimental results for MetaQA are shown in Table \ref{tbl:metaqa_results}. For 1-hop questions in MetaQA (which is identical to WikiMovies), our model is comparable to the state-of-the-art\footnote{6.7\% questions in Wikimovies are ambiguous, e.g. questions about movies with remakes without specifying years.}. For the other three settings, the performance of our re-implementation is slightly worse than the results reported in by original GRAFT-Net paper \cite{graftnet}; this is likely because we use a simpler retrieval module.

In the KB-only setting, PullNet shows a large improvement over the baseline models on 2-hop and 3-hop questions. For the text only setting, PullNet also improves on the baselines, and other prior models, by a large margin. Finally, PullNet also shows significant improvements over baselines in the text and incomplete-KB-plus-text settings.  We also see that PullNet is able to effectively combine KB and text information, as this setting also greatly outperforms the incomplete KB alone or the text alone.\footnote{Another way to perform QA in the incomplete KB plus text setting would be to ensemble two QA systems, one which uses the incomplete KB, and one which uses the text.  Although we do not make that comparison here, prior work \cite{graftnet} did show that GRAFT-Net outperforms such ``late fusion'' approaches.}

\subsubsection{WebQuestionsSP}
Table \ref{tbl:webqsp_results} presents the results on the WebQuestionsSP dataset. PullNet is comparable with GRAFT-Net on the complete KB only setting and slightly worse on the text only setting, but is consistently better than GRAFT-Net on the incomplete KB setting or the incomplete KB plus text setting.  It is also significantly better than the re-implemented GRAFT-Net, which uses the less highly-engineered retrieval module.

\input{webqsp_results.tex}

\subsubsection{Complex WebQuestions}

Complex WebQuestions contains multi-hop questions against FreeBase: intuitively, one would expect that single-shot retrieval of facts and text would not be able to always find efficient information to answer such questions.  Table \ref{tbl:complexwebq_dev_results} shows our results for Complex WebQuestions on the development set. An expected, PullNet shows significant improvement over GRAFT-Net and KV-Mem on all four settings. Once again we see some improvement when pairing the incomplete knowledge base with text, compared to using the incomplete knowledge base only or the text only.

\input{complexwebq_results.tex}

Researchers are only allowed limited submissions on the Complex WebQuestions test set, however, some results for the test set are shown in Table \ref{tbl:complexwebq_test_results}. Results with the text-only (Wikipedia) setting are comparable  to the dev set results (13.8\% instead of 13.1\%) as are results with the KB-only setting (45.9\% instead of 47.2\%).   

For completeness, we also compare to SplitQA \cite{complexwebq}.  SplitQA takes a multi-hop question and decomposes it into several simple sub-questions, and sends each of these to sub-questions to the Google search engine. After that, it applies a reading comprehension model to gather information from the web snippets returned by Google to find answers.  Using this same collection of Google snippets as the corpus, our model has 4.5\% lower hists-at-one than SplitQA.  However, the snippet corpus is arguably biased toward the SplitQA model, since it was collected specifically to support it.  We also note unlike SplitQA, PullNet relies only on open-source components and corpora. 

\subsection{Further Results}

\subsubsection{Retrieval Performance of PullNet}
\begin{figure}[htp]
    \centering
    \includegraphics[width=0.48\linewidth]{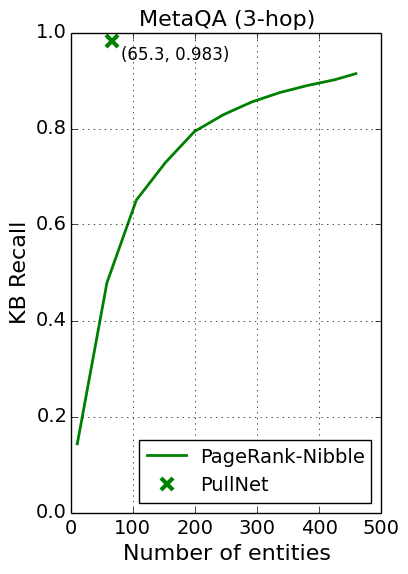}~
    \includegraphics[width=0.48\linewidth]{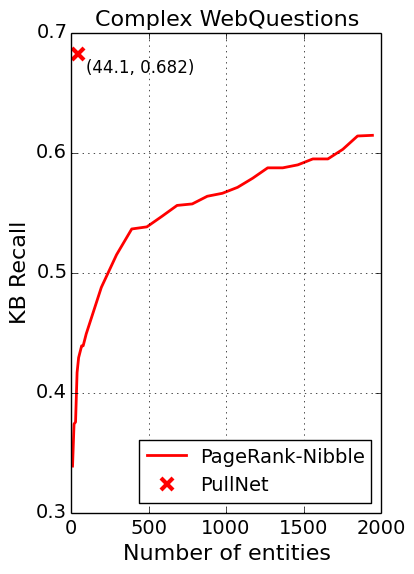}
    \caption{Recall of graphs retrieved by PageRank-Nibble compared with PullNet. \textbf{Left}: MetaQA (3-hop). \textbf{Right}: Complex WebQuestions.}
    \label{fig:kb_retrieval_pullnet_apr}
\vspace{-6pt}
\end{figure}

We compare the retrieval performance of PullNet and PageRank-Nibble on multi-hop questions with a complete KB, varying the number of entities retrieved by PageRank-Nibble. The results in Figure \ref{fig:kb_retrieval_pullnet_apr} show that PullNet retrieves far fewer entities but obtains higher recall. 

\begin{figure}[htp]
    \centering
    \includegraphics[width=0.48\linewidth]{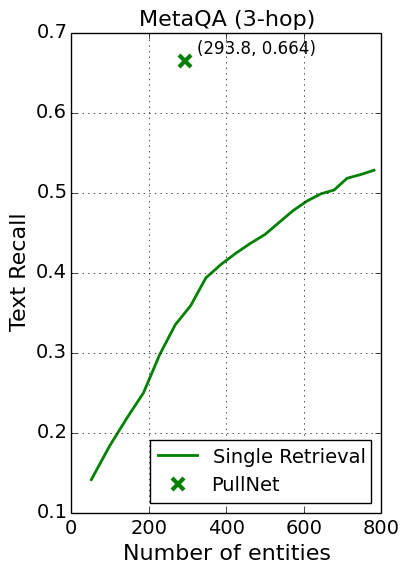}~
    \includegraphics[width=0.48\linewidth]{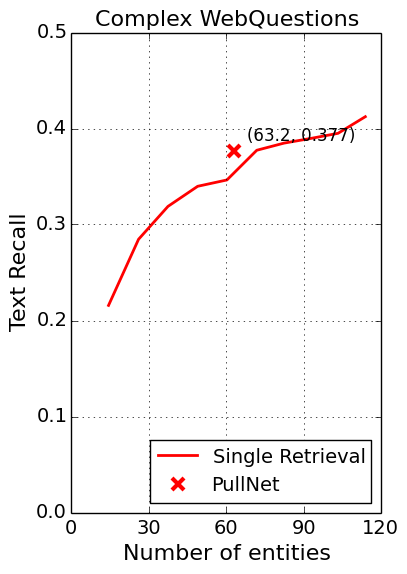}
    \caption{Recall of a single round of retrieval with Apache Lucene compared with PullNet. \textbf{Left}: MetaQA (3-hop). \textbf{Right}: Complex WebQuestions.}
    \label{fig:text_retrieval_pullnet_apr}
\end{figure}

We also further explored the effectiveness of iterative retrieval for multi-hop questions on a text corpus. The results are shown in Figure \ref{fig:text_retrieval_pullnet_apr}.  Again, PullNet with multiple iterations of retrieval, obtains higher recall than a single iteration of IDF-based retrieval on both the MetaQA (3-hop) and the Complex WebQuestions dataset. 

\begin{figure}[htp]
    \centering
    \includegraphics[width=0.9\linewidth]{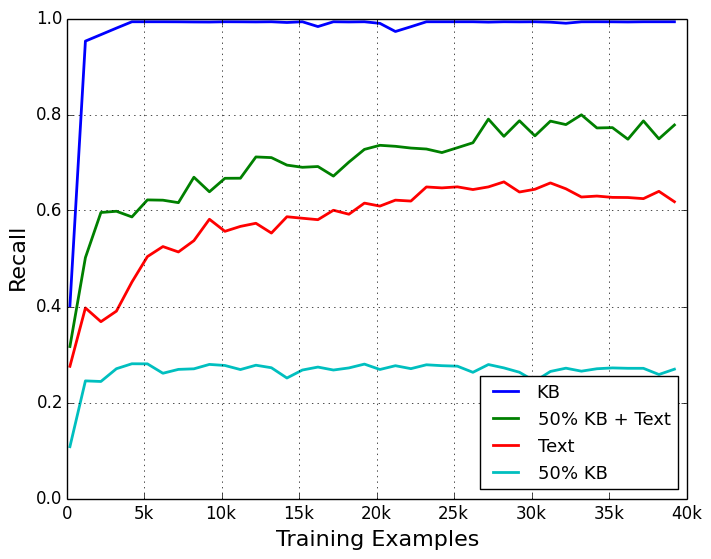}
    \caption{Recall of question subgraph on MetaQA 3-hop questions.}
    \label{fig:recall_iterations}
\end{figure}

Figure \ref{fig:recall_iterations} shows the recall of question subgraphs on MetaQA (3-hop) questions as training proceeds. Performance of the retrieval components of PullNet converges relatively quickly, with recall saturating after 10-20,000 examples (about 10-20\% of a single epoch).

\subsubsection{Training Efficiency}

We also analyze the training efficiency of PullNet. PullNet's algorithm is quite different from prior systems, since learning and retrieval are interleaved: in most prior systems, including GRAFT-Net, retrieval is performed only once, before learning. Intuitively, interleaving learning with the relatively slow operation of retrieval is potentially slower; on the other hand, PullNet's final question subgraph is smaller than GRAFT-Net, which makes learning potentially faster.

\begin{figure}[htp]
    \centering
    \includegraphics[width=0.9\linewidth]{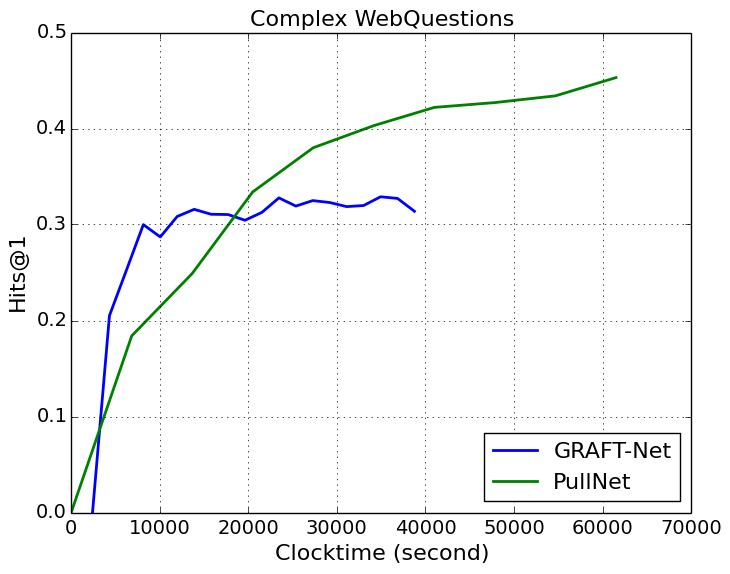}
    \caption{Performance of PullNet and GRAFT-Net under wall clock training time.}
    \label{fig:accuracy_clocktime}
\end{figure}

To study these issues, we plot the Hits@1 performance of learned model versus wall clock time in Figure \ref{fig:accuracy_clocktime}. This experiment is run on Complex WebQuestions in the KB-only setting, using one high-end GPU. To be fair to GRAFT-Net, we used a fast in-memory implementation of PageRank-Nibble (based on SciPy sparse matrices), which takes about 40 minutes to complete.  GRAFT-Net takes an average of 31.9 minutes per epoch, while PullNet is around four times slower, taking an average of 114 minutes per epoch.

As the graph shows, initially PullNet's performance is better, since GRAFT-Net cannot start learning until the preprocessing finishes.  GRAFT-Net's faster learning speed then allows it to dominates for some time.  GRAFT-Net reaches its peak performance in about 3.6 hours.  PullNet passes GRAFT-Net in hits-at-one after around 6 hours, or about 3 epochs for PullNet.

%% file: metaqa_results.tex
\begin{table*}[htp!]
\small
\begin{tabular}{lcccc|cccc|cccc}
\hline
                   & \multicolumn{4}{c}{MetaQA (1-hop) / wikimovies}                                                                             & \multicolumn{4}{c}{MetaQA (2-hop)}                                                                                        & \multicolumn{4}{c}{MetaQA (3-hop)}                                                                                          \\ \cline{2-13} 
                   & KB   & Text  & \begin{tabular}[c]{@{}c@{}}50\%\\ KB\end{tabular} & \begin{tabular}[c]{@{}c@{}}50\% KB \\ + Text\end{tabular} & KB   & Text  & \begin{tabular}[c]{@{}c@{}}50\%\\ KB\end{tabular} & \begin{tabular}[c]{@{}c@{}}50\%KB\\ + Text\end{tabular} & KB   & Text  & \begin{tabular}[c]{@{}c@{}}50\%\\ KB\end{tabular} & \begin{tabular}[c]{@{}c@{}}50\% KB\\  + Text\end{tabular} \\ \hline
KV-Mem*             & 96.2 & 75.4 & 63.6                                              & 75.7                                                     & 82.7 & 7.0    & 41.8                                              & 48.4                                                   & 48.9 & 19.5 & 37.6                                              & 35.2                                                     \\
GraftNet*    & 97.0   & 82.5 & 64                                                & 91.5                                                     & 94.8 & 36.2 & \textbf{52.6}                                              & 69.5                                                   & 77.7 & 40.2 & 59.2                                              & 66.4                                                     \\
PullNet (Ours)            & 97.0   & 84.4 & 65.1                                              & \textbf{92.4}                                                     & \textbf{99.9} & \textbf{81.0}   & 52.1                                              & \textbf{90.4}                                                   & \textbf{91.4} & \textbf{78.2} & \textbf{59.7}                                              & \textbf{85.2}                                                     \\ \hline \hline
KV-Mem   & 93.9 & 76.2 & --                                                & --                                                       & --   & --   & --                                                & --                                                     & --   & --   & --                                                & --                                                       \\
GraftNet & 96.8 & \textbf{86.6} & \textbf{68.0}                                                & \textbf{92.6}                                                     & --   & --   & --                                                & --                                                     & --   & --   & --                                                & --                                                       \\
VRN           & \textbf{97.5} & --   & --                                                & --                                                       & 89.9 & --   & --                                                & --                                                     & 62.5 & --   & --                                                & --                                                      \\ \hline
\end{tabular}
\caption{Hits@1 on MetaQA compared to baseline models. Number below the double line are from original papers: KV-Mem (KB) \cite{miller2016key}, KV-Mem (Text) \cite{watanabe2017question}, GraftNet \cite{graftnet}, and VRN \cite{metaqa}. * Reimplemented or different data.}
\label{tbl:metaqa_results}
\vspace{-12pt}
\end{table*}

%% file: webqsp_results.tex
\begin{table}[htp]
\small
\centering
\begin{tabular}{lcccc}
\hline
            & \multicolumn{4}{c}{WebQuestionsSP}                                                                                              \\ \cline{2-5} 
            & KB         & Text & \begin{tabular}[c]{@{}c@{}}50\%\\ KB\end{tabular} & \begin{tabular}[c]{@{}c@{}}50\% KB\\ + Text\end{tabular} \\ \hline
KV-Mem*       & 46.7       & 23.2 & 32.7                                              & 31.6                                                    \\
GraftNet* & 66.4       & 24.9 & 48.2                                              & 49.7                                                    \\
PullNet (Ours)     & \textbf{68.1}       & 24.8 & \textbf{50.3}                                              & \textbf{51.9}                                                    \\ \hline \hline
GraftNet    & \textbf{67.8}       & \textbf{25.3} & 47.7                                              & 49.9                                                    \\
NSM         & 69.0 (F1)  & --   & --                                                & --                                                     \\ \hline
\end{tabular}
\caption{Hits@1 on WebQuestionsSP compared to baseline models. Number below the double line are from original papers: GraftNet \cite{graftnet}, and NSM \cite{nsm} (which only reports F1 in their paper). * Reimplemented or different data.}
\label{tbl:webqsp_results}
\vspace{-6pt}
\end{table}

%% file: complexwebq_results.tex
\begin{table}[htp!]
\small
\centering
\begin{tabular}{lcccc}
\hline
         & \multicolumn{4}{c}{Complex WebQuestions (dev)}                                     \\ \cline{2-5} 
         & KB   & Text & \begin{tabular}[c]{@{}c@{}}50\%\\ KB\end{tabular} & \begin{tabular}[c]{@{}c@{}}50\% KB \\ + Text\end{tabular} \\ \hline
KV-Mem*       & 21.1 & 7.4  & 14.8    & 15.2                                                      \\
GraftNet* & 32.8 & 10.6 & 26.1    & 26.9                                                      \\
PullNet (Ours)  & \textbf{47.2} & \textbf{13.1} & \textbf{31.5}    & \textbf{33.7}                                                      \\ \hline
\end{tabular}
\caption{Hits@1 on Complex WebQuestions (dev) compared to baseline models. * Reimplemented or different data.}
\label{tbl:complexwebq_dev_results}
\vspace{-12pt}
\end{table}

\begin{table}[htp!]
\small
\centering
\begin{tabular}{lccc}
\hline
        & \multicolumn{3}{c}{Complex WebQuestions (test)}    \\ \cline{2-4} 
        & GoogleSnippet & Wikipedia & Freebase \\ \hline
SplitQA & \textbf{34.2}               & --             & --           \\
PullNet & 29.7               & 13.8           & 45.9         \\ \hline
\end{tabular}
\caption{Hits@1 on Complex WebQuestions (test).}
\label{tbl:complexwebq_test_results}
\vspace{-6pt}
\end{table}

%% file: 05_conclusion.tex
\section{Conclusions}
PullNet is a novel integrated QA framework for (1) learning what to retrieve from a KB and/or corpus and (2) reasoning with this heterogeneous to find the best answer.  Unlike prior work, PullNet uses an {iterative} process to construct a question-specific subgraph that contains information relevant to the question.  In each iteration, a graph-CNN is used to identify subgraph nodes that should be expanded using retrieval (or ``pull'') operations on the corpus and/or KB.  This iterative process makes it possible to retrieve a small graph that contains just the information relevant to a multi-hop question---a task that is in general difficult.

Experimentally PullNet improves over the prior state-of-the art for the setting in which questions are answered with a corpus plus an incomplete KB, or in settings in which questions need complex "multi-hop" reasoning.  Sometimes the performance improvements are dramatic: e.g., an improvement from 62.5\% hits-at-one to 91.4\% hits-at-one for 3-hop MetaQa with a KB, or improvements from 32.8\% to 47.2\% for Complex Web\-Questions with a KB.